# 6. Explainable artificial intelligence for Healthcare applications using Random Forest Classifier with LIME and SHAP

Mrutyunjaya Panda[1,*], Soumya Ranjan Mahanta[2]

[1]Associate Professor, Department of Computer Science, Utkal University, Vani Vihar, Bhubaneswar, Odisha-751004, India

mrutyunjaya74@gmail.com

[2] M. Tech. Student, Department of Computer Science, Utkal University, Vani Vihar, Bhubaneswar, Odisha-751004, India

dipusoumyaranjan019@gmail.com

*Corresponding Author

**Abstract:**

With the advances in computationally efficient artificial Intelligence (AI) techniques and its numerous applications in our every day's life, there is a pressing need to understand the computational details hidden in black box AI techniques such as: most popular machine learning and deep learning techniques; through more detailed explanations. The origin of explainable AI (xAI) is coined from these challenges and recently gained more attentions by the researchers by adding explainability comprehensively in traditional AI systems. This leads to develop an appropriate framework for successful applications of xAI in real life scenarios with respect to innovations, risk mitigation, ethical issues and logical values to the users. In this book chapter, an in-depth analysis of several xAI frameworks and methods including LIME (Local Interpretable Model-agnostic Explanations) and SHAP (SHapley Additive exPlanations) are provided. Random Forest Classifier as black box AI is used on a publicly available Diabetes symptoms dataset with LIME and SHAP for better interpretations. The results obtained are interesting in terms of transparency, valid and trustworthiness in diabetes disease prediction.

## 6.1. Introduction

In the recent past, applications of artificial intelligence techniques have seen exponential growth in every sphere of life, be it Computer vision, natural language processing, precision medicine, smart agriculture, or autonomous driving to name a few, despite its poor transparency and interpretability. The emerging deep learning architectures are posing even more complexity in interpreting and explaining the inner details of the black box approaches what they adopt. A diagrammatic representation of developed AI models based on the complexity is shown in Figure 6.1. From Fig. 6.1., it is quite evident that the recent most popular and widely used deep learning models are not only complex in design but also provide less explanations about its functionality in comparison to other existing AI approaches. As per European Union regulation 679 [1], the user has every right to not only understand about the usability of its data by AI models but also can challenge its predictions before accepting the solutions for necessary applications such as: precision healthcare, autonomous driving etc. This way, it is envisioned that xAI might be a proper one where AI with explainations may enhance trust in adherence to the several regulatory provisions and generate profits to the organizations. However, the challenge lies in making the xAI interpretable, transparent, trustworthy and complete [2]. The completeness comes from how accurately a system inner detail is being explained.

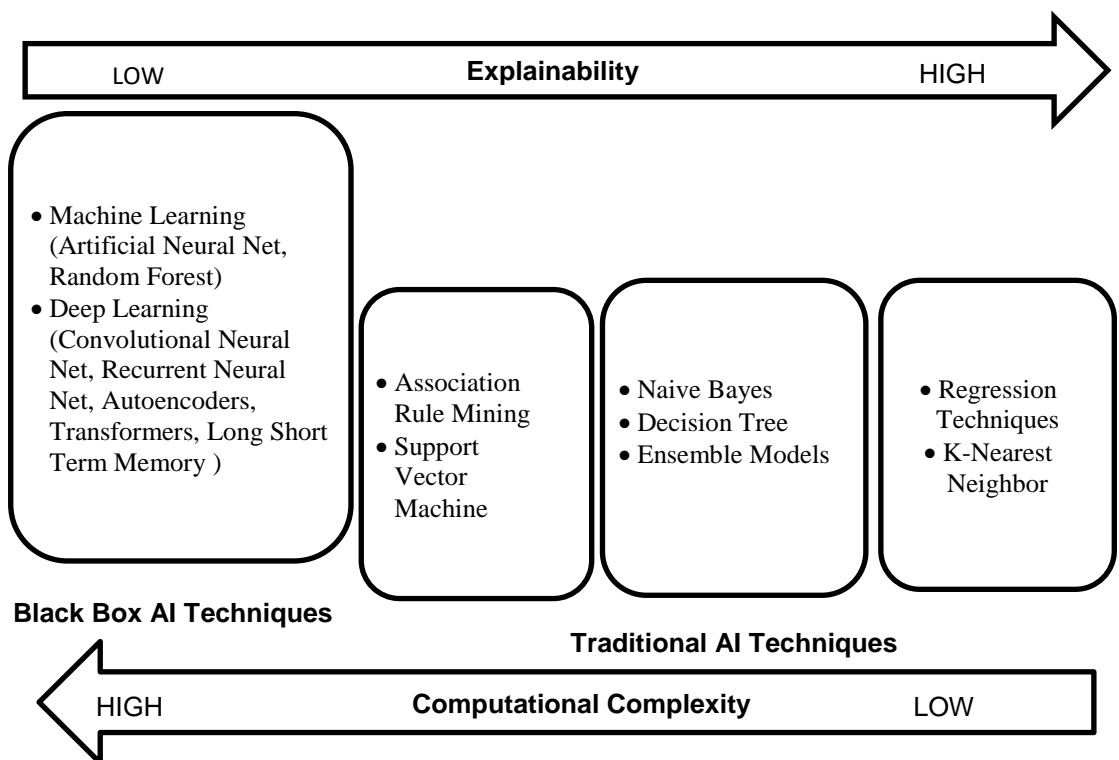

Figure 6.1. Classification of AI models in terms of Complexity and Explainability

**Motivation:**

Even though traditional AI methods are implemented to solve real life situations including healthcare informatics, self-driving cars, Self-driving drones, Chatbots, 6-G communications, Industrial IoT etc. since last several decades, but the "black box" view of these with very poor interpretability, explanability about the operations and low transparency poses a lack of trust in decision making process. Motivated by this, this chapter is intended to provide a study of several possible xAI frameworks for dealing with the complex decision making by means of user's trust on a medical dataset.

Looking into the recent developments in xAI, this chapter primarily focuses on the following:
- Which explainable AI methods are available?
- performance measures to evaluate the xAI methods
- Types of explanations with fairness and confidence
- How to choose amongst different explainable AI methods?
- Trust worthiness of xAI

The rest of the chapter is organized as follows: Section 6.2 presents the review of related literature stating current application scenario of xAI. Section 6.3 presents xAI and its framework followed by trust in xAI in Section 6.4. Section 6.5 discusses about the several xAI methods available in the literature. Experimental framework and results with discussions are presented in Section 6.6. Finally, conclusion and future work is highlighted in Section 6.7.

## 6.2. Related work

Mahbooba et al. [3] proposed to use the concept of xAI to improve trust management issues in intrusion detection system using decision tree model. The authors in [4] proposed a fine-grained mechanism based on segmentation and RoI pooling applied in Berkeley DeepDrive eXplanation (BDD-X) dataset and concluded that their approach outperforms the earlier methods with more interpretable visual explanations in autonomous driving system. It is discussed by the authors [5] that todays advanced bio-inspired algorithms which is used to tackle the real life situations, are considered
sometimes as a weak AI or narrow AI, as these are useful to address a specific application of interest whereas strong AI can be used for universal applications. Kim et al. [6] provided an investigative report of how explainability has been applied successfully in information systems till date and highlighted some quality criteria to judge the importance of xAI with future directions of research with pros and cons. Mane and Rao [7] applied deep learning methods on NSL-KDD dataset for network intrusion detection system with more explanations and then elaborated on how explanations generated from contrastive explanations helps in influencing the degree of attack prediction. In [8] the authors discussed about the acceptance of the customer in deployment of autonomous vehicles; highlighted the current regulatory provisions; different autonomous driving operations such as perception, localization, planning, control, and system management in detail. Kartikeya [9] presented two experimental framework to check whether transparency enhances model's trust or not; by using evaluation metrics and the other through the trust in subjective questionnaire; where after statistical significance test the author found that both provides results contrary to each other in measuring the models influence in terms of trust with transparency. Markus et al. [10] pointed out the xAI can provide trustworthy AI in biomedical informatics, however it needs further investigation to conclude its role in health care through verification and validation of data quality and external regulations. Hussain et al. [11] discussed the usefulness of xAI in autonomous cars from engineering perspective and discusses about its role in object detection, control strategies and decision making. The authors in [12] illustrated the applicability of xAI in several Internet of Things (IoT) enhancements including Internet of Medical Things (IoMT), Industrial IoT (IIoT), and Internet of City Things (IoCT) with added security and support from $6^{th}$ generation (6G) communication services. Mankodia et al. [13] applied various xAI methods for semantic object detection where deep learning models are used to segment and detect the road while the autonomous car is moving with maximum Intersection over Union (IoU) scores of 94.59% and 96.21%; accuracy of 97.61% and 97.86% for the training and testing dataset with ResNet-18. Javed et al. [14] presented a survey report on usefulness of XAI in developing smart cities which is considered to be a noble idea and highlighted its research directions on system architectures. Renda et al. [15] proposes to use the concept federated learning of xAI in 6G communications for autonomous driving in intelligent transportation system with benefits of trust, quality of experience and privacy preserving management of the system design. Kim and Joe [16] applied xAI in autonomous driving using convolutional neural network (CNN) by explaining the differences in output values obtained in final hidden layer of CNN by using image sensitivity analysis on image data and finally concludes that xAI could categorize the images with high accuracy. Madanu et al. [17] discusses about the role of AI in pain modelling using machine learning and deep learning techniques and then advocates that the xAI models might add more detailed diagnostic analysis with incorporations of explainability to a certain illness with its root cause analysis.
For this purpose, pre-existing models were applied on the Google Jigsaw dataset to get the best prediction accuracy, and explainable methods such as LIME (local interpretable model—agnostic explanations) were applied to the HateXplain dataset [18]. Variants of BERT (bidirectional encoder representations from transformers) model such as BERT + ANN (artificial neural network) and BERT + MLP (multilayer perceptron) were created to achieve a good performance in terms of explainability using the ERASER (evaluating rationales and simple English reasoning) benchmark [19].

### 6.3. Explainable AI and its frameworks

Several Explainable AI (xAI) frameworks are discussed in this section to address the following:
- Explainability
- Interpretability
- Transparency
- Trustworthiness
- Evaluation criteria

#### 6.3.1. Explainability

Transparency aims at getting the inner working details of the AI model being developed, so that a sense of human understanding is achieved and mutual trust can be ensured while using the model. In this aspect, it is observed that sometimes researchers finds it hard to clearly define the explainability and finally trapped into the definitions of interpretability and/or into transparency [20] ].

#### 6.3.2. Transparency

There are three dimensions in transparency for human understanding about the inherent details of the AI model being considered for applications under study. They are:

- **Simulation Ability**: This is the first level of transparency where the AI model's ability to be simple and compact as well so that a human can be able to simulate the model with ease, but not at the expense of efficient decision making. Artificial neural network with no hidden layers might fall in this category.
- **Decomposition of the model**: This is the second level of transparency where the AI model can be divided into several parts in terms of Input, mathematical computations and setting of the model parameters and then a detailed explanation of each individual parts are obtained for better understanding of the human. In this level, there are only a few AI model available those who could satisfy this criteria.
- **Transparency in Al model procedure**: In this third transparency, one can envisage of getting the better understanding of the procedures being undertaken to develop the AI model and the way how the output is obtained. One such example is k-Means clustering algorithm, where the distance based similarity criteria is well mentioned so that the samples with high similarity can be placed in same cluster. In contrast, if we see the deep neural network models, loss function used seems to be elusive and the objective to train the Neural network model is done through approximate reasoning, hence lack of transparency is observed to understand the inner details while reaching output solution. Hence, a AI model with mathematical analysis falls into this transparency level [21].

Looking all these discussions, the AI models are classified as either transparent or black box /opaque model. Examples of Decision Tree, Linear regression, K-Nearest

Neighbour (KNN) and fuzzy classifiers etc. are falls under transparent model whereas all variations of neural networks except a single perceptron neural network, Random Forest, Support Vector Machine (SVM) and Microsoft's recent fiasco with the experimental Tay Twitter Chatbot etc. fall into opaque/non-transparent models.

It is also to be noted here that opaque AI models are more effective than their counterparts due to no reasoning constraints but with higher risk. So, there is a trade-off between these choices by the companies or individuals looking into their specific requirements.

### 6.3.3. Quality Criteria

The evaluation of AI model is done through quality of information they generate for effective and efficient decision making process. There are two aspects of these quality criteria for model evaluations such as: Explanation aspects and the model aspects. In explanation aspects, the following quality criteria are often used to make the AI model explainable.

- **Comprehensibility:** Comprehensibility refers to the ability of AI model to represent its learning outcomes in a human understandable manner [22].

- **Accessibility:** Accessibility intends to involve the end-users in improvement of the AI model without having any deep understanding of the AI programming [23].

- **Fidelity:** Fidelity discusses how accurately the explanation method justifies the underlying model from which the difference between the users could be able to measure the difference in its descriptive model accuracy with that of system generated ones [24].

- **Identity, Separability and Novelty**: where the explanations between different instances are compared with identical instances should have identical explanations; non-identical ones should have separate explanations and the instance should not have come from a region in instance space too far from the training data.

Apart from these quality criteria, the following ones are used for obtaining explanations from the model aspects. In this, the output from the users' aspect or mental model is fed back to the system via the criteria appropriate trust and reliance[25].

- Accuracy is one of the model evaluation criteria through which one can evaluate the level of model's correctness in comparison to the actual target
- Fairness which is also understood as opposite to model biasness which presents the degree of error patterns present in the AI model
- Reliability presents soundness of the model to the user for a specific application.

### 6.3.4. Types of Explainability techniques

The explainability techniques in xAI are basically of two types: Global and Local. In global explainability, model explanations are made in general with its generic operations whereas in local ones, the explanation comes for every single data with model reasoning and appropriate rules through which a decision is obtained.

- **Types of explanations:** In general, there are two categories of explanations: intrinsic and post hoc explanations (or interpretability). In the case of intrinsic explanations, simple AI models that are interpretable due to its structural simplicity such as sparse linear models and decision trees are considered whereas interpretations are obtained after the model is trained in case of post hoc explanations. In this chapter, our focus is on post hoc explanations. Examples of post hoc explanation methods for image and time series datasets are factual, semi-factual and counterfactual methods.

**Post-hoc explanations:** Post-hoc explanations are traditionally viewed as explanations or justification of decision making process by example where some factual (or nearest neighbor) instance is created to justify some target query [26]. However, there are several other explanatory possibilities are available based on the type of application using the explanations. Let us consider an opaque classifier with student dataset with student placement attribute available in a tabular format to obtain the decision whether a student will get placement or not. Now, supposing that you did not get selected in campus placement and for which you asked explanations about the non-selection. Then, the AI model could return answer to your query explanation through a factual example-based explanation as 'you are not selected because your interview was similar to another student who also did not get placement". Alternately, the AI model could give you some counterfactual explanations 'if you would have score better in interview like the Student-X who got selected for placement". At last, there is also a chance of getting semi-factual explanations with reply "even if you performed well in interview, you would still not have the positive attitude of Student-X who got selected".  It is customary to say that factual post-hoc example based explanations are most widely used by the researchers' as in case based reasoning classifiers and these alternate explanations are considered to be non-trivial, but it is observed that there is a growing demand in using counterfactual [27] and semi factual explanation methods [28].

### 6.4. Trust in xAI

Trust can be considered as an abductive speculation which seems to be the "best hypothesis" to measure the trustworthiness in AI. As this consideration is purely based on the intelligibility and anticipated predictability of the AI model, hence is undoubtedly fallible [29]. To this, one can envisage that what a computer scientist understands about the implications of computer assertions, the others may not and thereby ask for justification to believe that the computer assertions are valid and trustworthy. Looking into this, trust in AI is defined in several ways:
- Absolute Trusting, where the novice user considers the computer assertions are completely valid and trustworthiness in all conditions.
- Contingent Trusting is one where the customer accepts the computer assertions as valid and trustworthy under some specific conditions.
- Progressive Trusting on the other hand takes users experience on given computer assertions over a period of time to decide whether it is valid and trustworthy or not.
- Digressive Trusting by user finally considers some part of the computer assertions to conclude whether they are valid and trustworthy based on the users' experiences over time.

The trust in xAI through clarification and explanation with fairness in AI model development process can be seen from Figure 6.2.

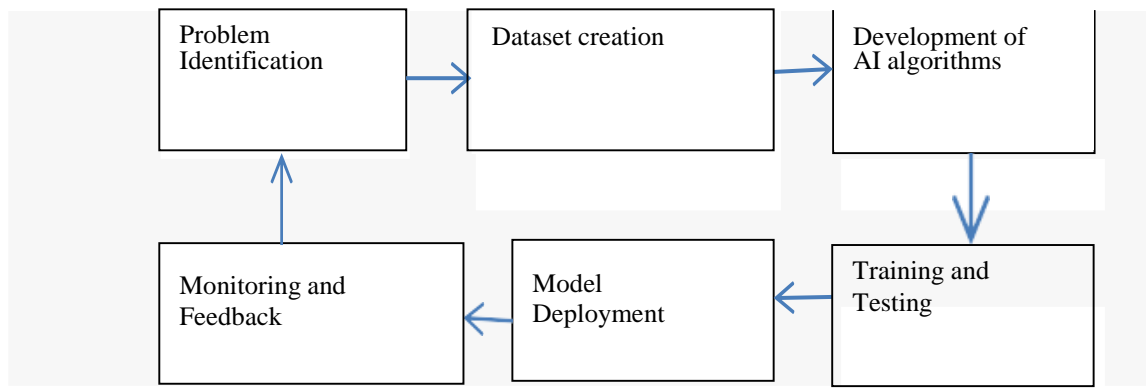

**Figure 6. 2: Explainability and fairness for measuring Trust in xAI**

In problem identification, the aim is to find whether the algorithm under study presents some ethical solutions to the problem at hand or not. Accordingly dataset is to be created by checking the representations from several groups and find out whether there is any bias available in the data or not. If any bias is present in either in class label or infeatures of the dataset used, steps to be taken to remove or minimize the bias for bettermodel predictability. Then, in the development of AI algorithms, fairness to be included as a constraint to the objective function, so that transparency of selecting the algorithmis well understood by the user. Next, dataset is slit into training and testing set. During training, model is properly trained and then checked with possible fairness matrix during testing phase. In deployment phase, the model is verified for its intended uses and its effectiveness in diverse applications. Finally, model is monitored for any unfairmeans if at all adopted during the whole process with obtained feedbacks and the process continues for further improvement.

### 6.5. xAI Methods

xAI methods or interpretation tools are categorized broadly into two types as: model agnostic and model-specific ones, depending upon whether the explanations are required for all types of models or tailor made for certain model structure respectively. In model-agnostic xAI methods, even though inner details and structure of the models cannot be accessed but it will provide an understanding for any of the AI model being used earlier whereas in the model specific methods, the interpretation depends on the working capabilities of the specific model being considered. Further, some xAI methods are considered to be global where the model tries to approximate the function underlying the model in whole space whereas in local methods, a small representative of the whole space is chosen and then approximation of that specific part is chosen for explanations. It is also seen that xAI methods provides interpretations based on feature importance, accordingly a rightful way is envisaged through correlation so as to find the important one among variables and then that is used to provide explanations. Global explanations provide interpretations to non-data scientists about what the model uses to make predictions. For example, in a customer churn prediction by a telecom company, the marketing team shall explore about the several important features those contribute in predicting the customer churn. On the other hand, local explanations identifies the most important dimension of a single input to predict its output as can be seen in medical diagnosis using deep neural network.
Based on the above discussions, the following are some of the local and global model-agnostic xAI methods available in the literature to obtain explanations: Partial Dependence Plot (PDP), Accumulated Local Effects (ALE), Individual Conditional Expectation (ICE), Local Interpretable Model-agnostic Explanations (LIME), SHapley Additive exPlanations (SHAP), Generalized additive model (GAM), moDel Agnostic Language for Exploration and eXplanation (Dalex), Contextual Importance and Utility (CIU) etc. . Out of all these, PDP, ALE, CIU and ICE are considered to be global interpretable model-agnostic explanations whereas LIME, SHAP, Dalex and GAM. The brief discussion about these methods are provided below.

### 6.5.1. Partial Dependence Plot (PDP)

Partial Dependence Plot (PDP) is model-agnostic and global interpretable method in xAI where the panoramic explanations about the influence of features on the target variable ones in the frame of whole dataset. It is observed that PDP has very nominal effect on the complement features (i.e. other than the intended ones) while prediction using AI models [30].

Further, in PDP, one can find whether the relationship between the target variable and an input feature is linear, monotonic or more complex. The formal definition of a PDP function for regression operation can be represented as presented in Eqn (1).

$$f_s'(x_s) = E_{X_c}\left[f_s'(x_s, X_c)\right] = \int f_s'(x_s, X_c) dP(X_c) \quad \text{---(1)}$$

where $x_s$ are the features for which PDP plot to be obtained and $X_c$ are the complement features present in the AI model and $f_s'$ are random variables. In this, set S contains one or two features which are used to understand the outcome of the prediction. The full feature space can be formed by combining the feature vectors $x_s$ and $X_c$. PDP works by disparage the output of the black box model over the features distribution in set C, through which PDP function can show the association between features of interest from set S and the predicted output. On the other hand, if we disparage the complement features, then the obtained PDP function hang on between features in set S and the other features from set C , not in set S are included. The PDP function $f_s$' can be calculated by a popular method known as Monte Carlo method by contriving means in the training dataset. Even though, PDP is straight forward in displaying the linkage between a feature and the target, the independence assumption between the features of interest and the other features has become a major issue. Without meeting this assumption, interpretable and reliable PDP may not be feasible. At the same time, the drawback of PDP is that it can work well up to two features to have a comprehensible plot. In Python, Sklearn Inspection module may be used for getting PDP display for better interpretation of the AI model. Another such plotting method similar to PDP is ICE (Individual Conditional Expectation) to envision and inspect the interaction between the set of interest features with the target output.

### 6.5.2. Accumulated Local Effects (ALE)

Accumulated Local Effect (ALE) overcomes the drawbacks of PDP and ICE plots in estimating feature effects with much accuracy. However, approximation through ALE is not away from limitations. The drawbacks of ALE lie in its proneness to out-of- distribution (OOD) sampling for small training dataset and its inability to scale well for high dimensional input dataset [31].

### 6.5.3. Local Interpretable Model-agnostic Explanations (LIME)

Lime is short for Local Interpretable Model-Agnostic Explanations. Each part of the name reflects something that we desire in explanations. Local refers to local fidelity – i.e., we want the explanation to really reflect the behaviour of the classifier "around" the instance being predicted.

### 6.5.4. SHapley Additive exPlanations (SHAP)

SHapley Additive exPlanations (SHAP) dependence plot is a local, model-agnostic xAI framework which is used for image, tabular and text datasets for interpretability with Shaply values, a concept inspired from cooperative Game Theory. The prime objective of the SHAP is to provide explanations in AI models prediction by calculating the contribution of each individual feature to the output prediction. In this, feature values of a data sample denoted as players in coalition and the Shapley values is calculated as the mean marginal contributions a feature value across all possible coalitions [32]. The features having large shapley values considered to more important and the plotting of features is done based on their decreasing order of importance. SHAP dependence plot has advantages of fast implementation in tree based models by removing the slow computation barriers of shapley values. In python, SHAp package is available for xAI analysis. It is also observed that SHAP has some proneness while computing Shapley values for a lot of data samples. It's also introduces biases through some evil data scientists by introducing intentionally designed misleading explanations.

### 6.5.5. Generalized additive model (GAM)

A Generalised Additive Model (GAM), an extension of the multiple linear model is extremely flexible in choosing any kind of non-linear and linear regression models that are more appropriate to different types of output predictions. In one hand, GAM uses additivity property for explaining the contribution of each individual predictors while fixing other predictors, on the other hand this also poses some constraints in correlating the predictors' non-linear explanations automatically [33].

### 6.5.6. Contextual Importance and Utility (CIU)

Contextual Importance and Utility (CIU) arithmetic uses the concepts of Multi- Attribute Utility

Theory, with novel concept of contextual influence makes it possible to compare CIU directly with so-called additive feature attribution (AFA) methods for model-agnostic outcome explanation. It is to be noted here that the 'influence' conceptused by AFA methods is deficient for outcome explanation even for simple models. Further, CIU generates faithful explanations using contextual importance (CI) and contextual utility (CU) for outcome predictions where influence-based methods fail [34].

### 6.6. Experiments and Results

#### 6.6.1. Experimental framework

The experimental framework of the proposed approach is shown in Figure 6.3.

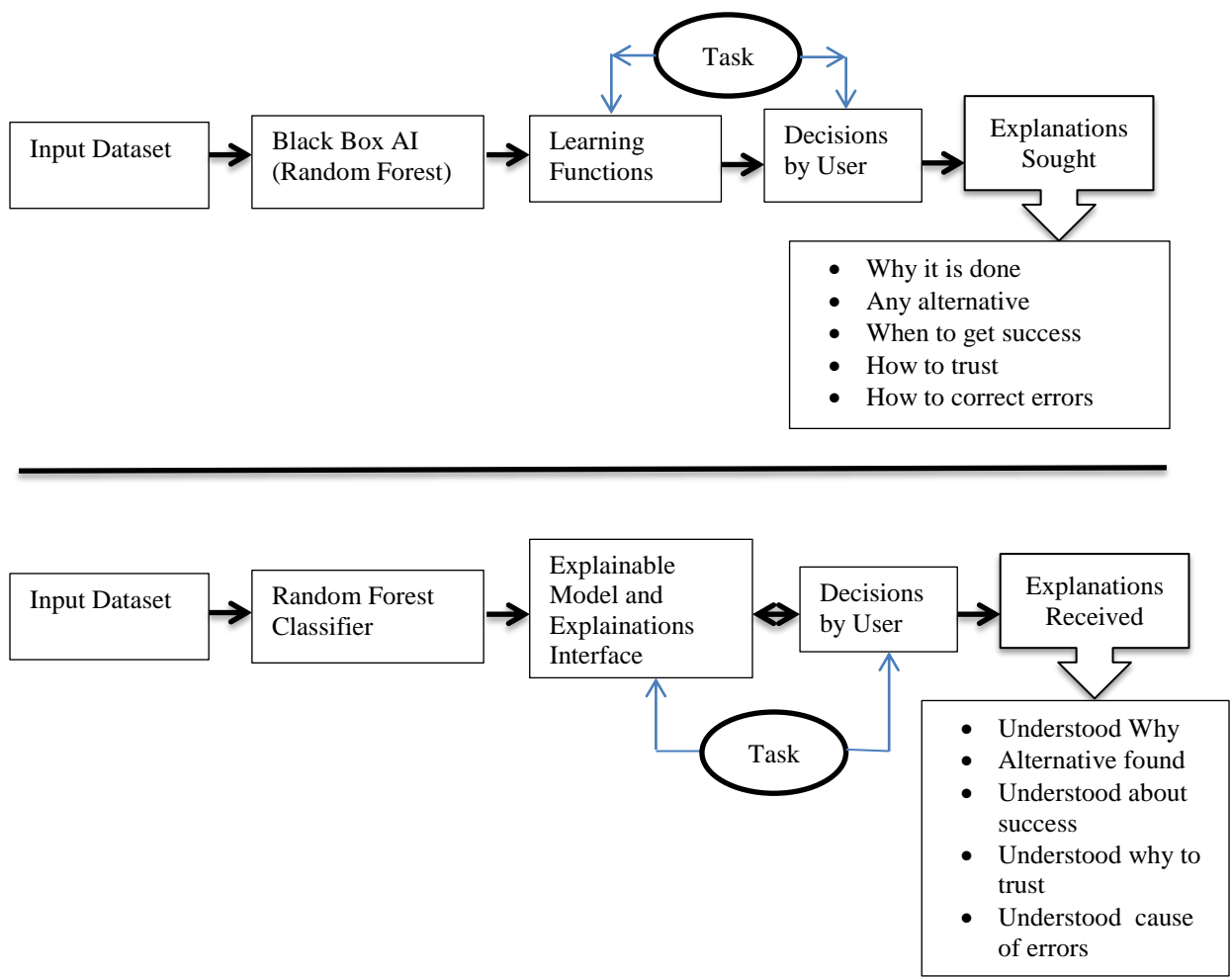

**Figure 6.3: The Proposed xAI framework**

From Figure 6.3, top part shows the black box model of AI where the user has lots of question in its mind to understand the inner details of the process through which a decision has come out about experiments, but in the xAI as shown in bottom part of Figure 6.3, it is seen a happy user who has understood the details about how and why thedecision is made with full satisfaction. From Figure 6.3, in xAI framework, at first the input dataset is collected and all pre- processing that are necessary to make a quality data is performed. Then the modified data is applied to new machine learning model such as Deep neural network model and the predictions are obtained. The predicted outcome from deep neural network is passed through any xAI model (LIME, SHAP etc.) to obtain the detailed explanations about the decisions being taken in the process. Finally, an explanation interface like: temeons xAI, a platform-agnostic freestanding upshot is used to integrate the xAI model with any arbitrator or user. By this way, xAI pursues a paradigm shift towards more user- and society-friendly AI categorization without compromising efficacy.

### 6.6.2. Experimental Results and Discussion

All the experiments are conducted in a HP Pavilion Intel Core i5 PC with 1 TB HDD, 8GB RAM using Python in Jupyter Notebook. Publicly available Diabetes dataset collected from Kaggle and Random Forest as black box AI method is used for development of the prediction of the diabetic disease. Finally, two surrogate xAI model such as: LIMA and SHAP are used for explanations to the inner details of the AI model in decision making process.

At first, the black box model performance by the random forest algorithm in terms of precision, recall and f1 score is presented in Figure 6.4. Here, 0 indicates the person having no diabetes and 1 for with diabetes.

```
              precision    recall  f1-score   support

           0       0.79      0.85      0.82       150
           1       0.68      0.58      0.63        81

    accuracy                           0.76       231
   macro avg       0.74      0.72      0.72       231
weighted avg       0.75      0.76      0.75       231
```

**Figure 6.4: Prediction performance of the Random Forest Algorithm**

Now, in order to add explanations to the model developed, global interpretation of the feature importance in the dataset is presented in Figure 6.5 using SHAP explainer.

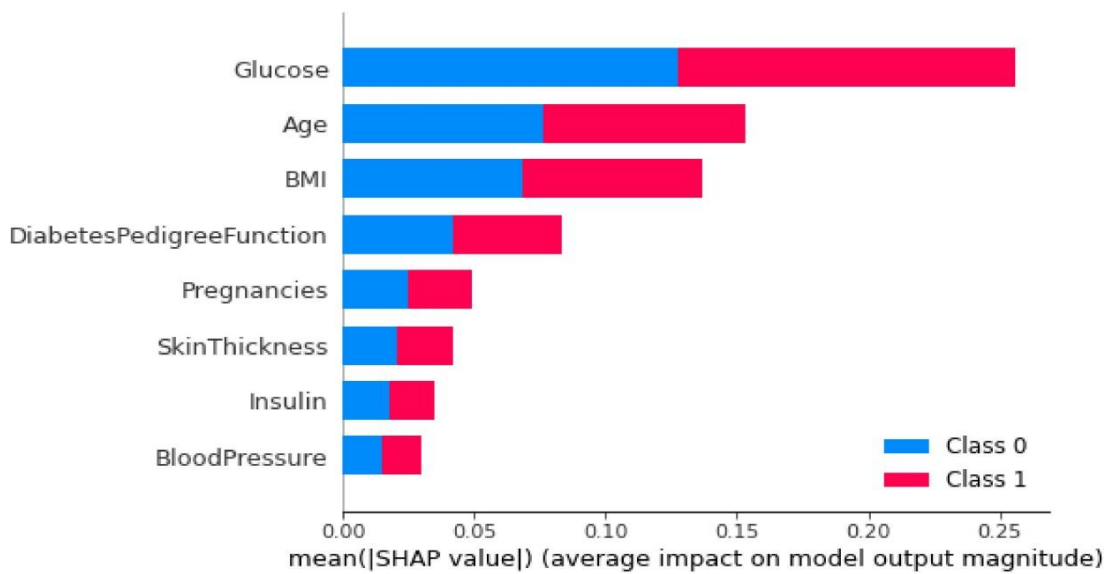

**Figure 6.5: Global interpretation with Feature importance plot using SHAP**

From Figure 6.5, one can observe that all the features equally contributing to both the class (Class 0 without diabetes and Class 1 is having diabetes) as red and blue are evenly present in all the features. Further, Glucose is having highest significance in the disease followed by age and BMI index. Similarly, another way to represent the feature importance map is shown in Figure 6.6 using the SHAP explainer, with its SHAP values. Here, the red dot indicates more significance in comparison to blue dot. For Glucose, more red dots in positive x-axis implies positive impact on the model predictions, whereas the blue dots in positive x-axis shows the low probability being detected as diabetes.

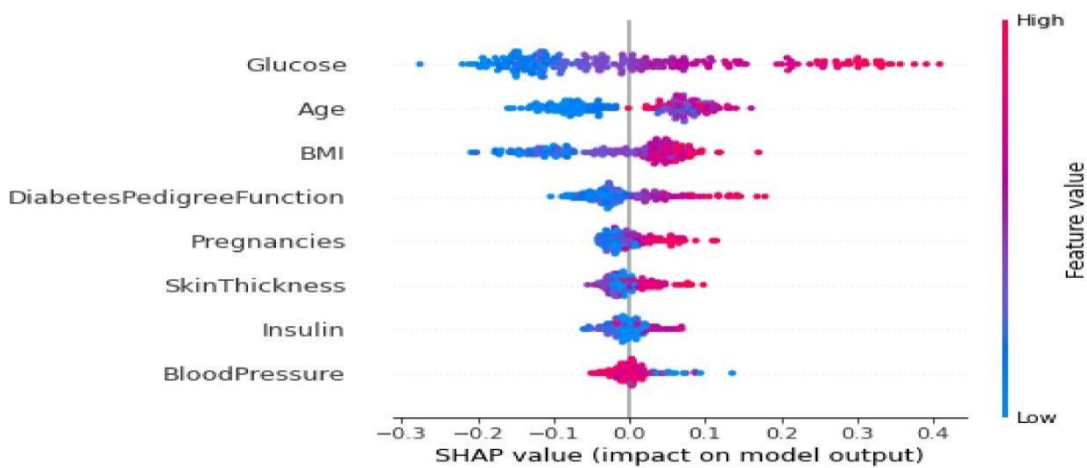

**Figure 6.6: Shapley values with explanations for model output**

Finally, SHAP dependency plot for age is used for more detailed explanations, as age provides confusion about the feature significance for its equal distributions of the points on both sides of the axes, which is shown in Figure 6.7. From Figure 6.7, one can interprete clearly that patients with age less than 30 is not having diabetes whereas above the age30, mostly found with diabetes.

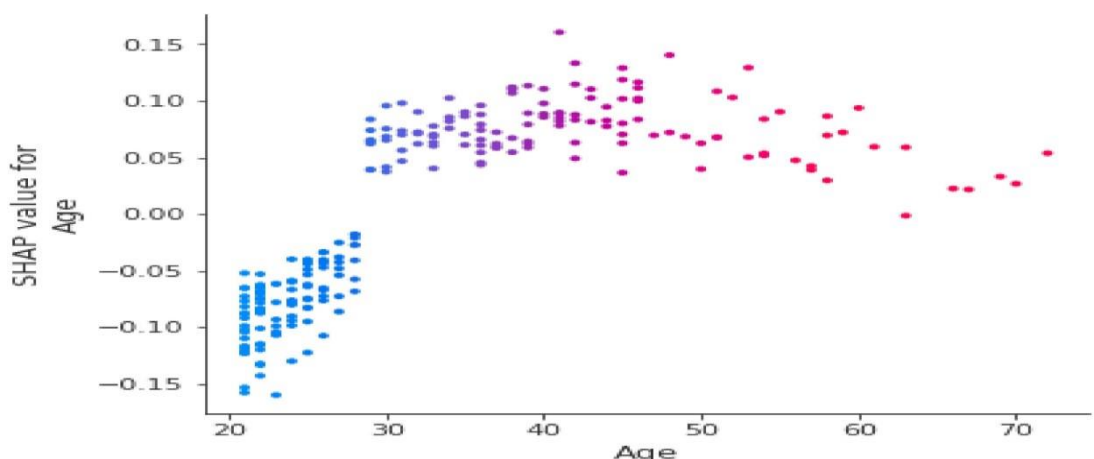

**Figure 6.7: SHAP dependency plot for age in diabetes dataset**

Now, we used LIME as local interpretable xAI method to understand the explanability of the prediction model outcome and feature relevance analysis. In this, two class level with value 1 and 0 are represented as having diabetes or no diabetes respectively. Only one sample from testing dataset is used for explanations. Next, LIME is used to show the explanations for the model developed using Random Forest algorithm on the diabetes dataset and the results obtained are presented in Figure 6.8. Figure 6.8 reveals that the single patient sample with an age of 37 and glucose level of 154 has 69% model predictability of having diabetes symptoms with following explanations: low insulin (126), low body mass index (BMI) (31.30), low Skin Thickness (29), pregnancies (4) and high blood pressure (72) in comparison to their respective thresholds as mentioned left side of the detailed feature values.

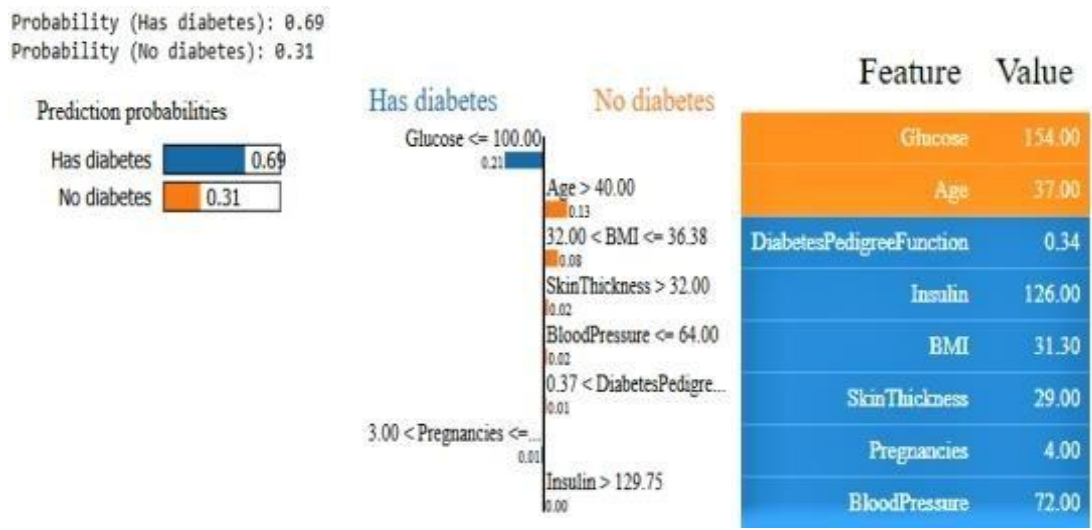

**Figure 6.8: Explanation of Random Forest model using LIME for persons with diabetes**

### 6.7. Conclusions

This chapter discusses the basics of explainable AI with several types of explanations with example. Further, it is aimed at discussing the xAI framework and methods available for post-hoc interpretation of the black box AI model to enhance the understanding of the predicted model for more transparency and acceptance. Next, trustworthiness of AI is discussed with various definition of trust in model development. Out of several xAI methods available, random forest machine learning model is used on a publicly available diabetes dataset for prediction of the diabetes and then SHAP and LIME methods are used for its explanability. Feature significance and dependence plots for SHAP as a global interpretability method along with LIME, as a local interpretability method are discussed, which added more insights to the decision making process. In future, the aim is to explore further possibilities with other black box AI as well as explanation techniques to have better explanations in decision making with trustworthiness.